\documentclass[runningheads]{llncs}
\usepackage{eccv}
\usepackage{eccvabbrv}
\usepackage{graphicx}
\usepackage{booktabs}
\usepackage{amsmath,amssymb}
\usepackage[accsupp]{axessibility}  
\usepackage{array}
\usepackage[table]{xcolor}
\usepackage{multirow}
\usepackage{adjustbox}
\usepackage{fix-cm}
\usepackage{comment}
\usepackage{makecell}

\newlength{\thickarrayrulewidth}
\makeatletter
\def\thickhline{%
  \noalign{\ifnum0=`}\fi\hrule \@height \thickarrayrulewidth \futurelet
  \reserved@a\@xthickhline}
\def\@xthickhline{\ifx\reserved@a\thickhline
    \vskip\doublerulesep
    \vskip-\thickarrayrulewidth
  \fi
  \ifnum0=`{\fi}}
\makeatother
\usepackage{hyperref}
\usepackage{orcidlink}

\begin{document}
\emergencystretch=2em
\setlength{\tabcolsep}{4.5pt}

\title{Reuse Before You Retrieve: Diagnosing Headroom and Complementarity for Test-Time Augmentation of Embodied Multimodal Policies}
\titlerunning{Reuse Before You Retrieve}

\author{Yuhwan Jeong\orcidlink{0009-0002-0279-146X} \and
Kuk-Jin Yoon\orcidlink{0000-0002-1634-2756}}
\authorrunning{Jeong et al.}
\institute{KAIST, Visual Intelligence Lab.\\
\email{\{jeongyh98, kjyoon\}@kaist.ac.kr}}
\maketitle

\begin{abstract}
Frozen vision-language-action (VLA) policies are increasingly improved at test time by sampling additional policy behaviors or introducing external demonstrations. Yet there is little guidance for deciding which intervention a deployed policy actually needs. Additional sampling is useful only when better behavior already exists within the policy's stochastic rollouts and can be identified, whereas retrieval is most useful when the relevant action prior is not reliably represented by the policy. We study this decision through two measurable factors, recoverable headroom and retrieval complementarity, which characterize how much useful behavior is already available to recover and whether an external action prior fills a measurable gap. We evaluate an episode-level retry selector under retryable or parallel execution, together with retrieval across multiple frozen VLA policies and environments. The selector consistently recovers substantial latent capability across all tested VLA backbones on LIBERO, with gains of up to 21.0 success-rate points that closely track recoverable headroom. It also transfers to a different robot and simulator and remains effective under degraded observations, while experiments with autoregressive OpenVLA illustrate the distinction between available headroom and the ability to rank candidate rollouts. Retrieval behaves differently, improving the policy with the largest measured action-prior gap and providing further gains when combined with selection. Together, these results provide an empirical basis for characterizing test-time augmentation opportunities by separating capability that can be recovered from the frozen policy from behavioral priors that may need to be introduced externally.
\end{abstract}

\section{Introduction}
Vision-language-action (VLA) models~\cite{black2024pi0,intelligence2025pi05,shukor2025smolvla,kim2024openvla} have shown that a single pretrained policy can follow language instructions and perform a broad range of manipulation tasks. However, even strong frozen policies remain imperfect when deployed under new tasks, environments, or observation conditions. Retraining such models for every deployment setting is costly, which has motivated a growing set of methods that improve policies at test time without updating their weights. These methods retrieve external demonstrations or memories~\cite{du2023behavior,memmel2024strap,li2025mapvla,park2026retrieve} or use additional test-time computation to sample and select among multiple policy outputs~\cite{snell2024scaling,kwok2025robomonkey,jang2025mgselect,dai2025rover}.

A recurring difficulty is that test-time interventions do not help consistently. The same retrieval mechanism can improve one policy while having little effect on another. Additional sampling is useful only when the policy already produces better behaviors that can be identified. This creates a practical decision problem for augmenting a frozen policy at test time. Should we attempt to recover better behavior already present in the policy's stochastic executions, or introduce external demonstrations to change the behavior the policy produces? Applying either intervention without understanding the underlying opportunity can waste resources or introduce a prior that the policy already represents. We therefore ask whether measurable properties of frozen policy behavior can distinguish the opportunities targeted by these forms of test-time augmentation.

A frozen policy can fail for fundamentally different reasons. It may already produce successful behavior under some stochastic rollouts but fail to do so consistently, making repeated execution and selection useful when retries or parallel systems are available. Alternatively, repeated rollouts may reproduce the same systematic failure, in which case additional sampling provides little opportunity and an external behavioral prior may be needed. These cases can have similar single-rollout performance but require different test-time interventions. A useful diagnostic should therefore distinguish failures that can be addressed by recovering existing capability from those that may benefit from introducing a missing behavioral prior.

We study this decision through two factors, \textbf{recoverable headroom} and \textbf{retrieval complementarity}. Recoverable headroom captures the opportunity already present within the policy's stochastic behavior. A policy may produce successful behavior across repeated rollouts but fail to realize it consistently in a single execution. We quantify this opportunity by the gap between pass@1 and pass@$N$. For retrieval, we operationalize complementarity by asking whether the retrieved behavior supplies an action prior that the frozen policy does not reliably produce on its own. Together, these measurements distinguish opportunities to recover capability already present in the policy from those that may benefit from introducing an external prior.

We examine these opportunities using two representative test-time interventions across frozen VLA policies. First, when repeated or parallel execution is available, we use an \emph{episode-level retry selector} that executes multiple stochastic rollouts from the same initial condition and selects the trajectory that remains closest to a demonstration manifold. This setting allows the selector to exploit stochastic capability already exposed by the frozen policy without modifying its parameters. Second, we use \emph{event-schema retrieval} to warm-start action generation with a retrieved demonstration chunk. To characterize the corresponding opportunities independently of downstream gains, we estimate recoverable headroom directly from repeated stochastic rollouts and measure how closely each policy can reproduce demonstrated actions. We evaluate three flow-matching VLA backbones, $\pi_0$, SmolVLA, and $\pi_{0.5}$, on LIBERO, and further test the main findings across SimplerEnv-Bridge, autoregressive OpenVLA, and degraded observations.

Our experiments reveal a consistent pattern. The policy's own stochastic rollouts contain substantial recoverable capability across all three LIBERO backbones, and the episode-level selector improves every backbone and suite combination, with gains of up to $21.0$ success-rate points that closely track recoverable headroom. The same selector also transfers to a different robot and simulator and remains effective under degraded observations. The OpenVLA experiment reveals a complementary limitation, substantial oracle headroom can remain difficult to recover when the available ranking signal is weak. Retrieval behaves differently. It consistently improves $\pi_0$, which exhibits the largest measured gap from the demonstrated action prior, while providing only small and inconsistent gains for SmolVLA and $\pi_{0.5}$. When both opportunities are present, retrieval and selection provide additional gains when combined. These findings motivate distinguishing the source of available test-time improvement when augmenting a frozen policy.

Our contributions are
\begin{itemize}
\item We introduce recoverable headroom and retrieval complementarity as measurable diagnostics for distinguishing between test-time opportunities that reuse a frozen policy's existing behavior and those that may benefit from an external action prior.
\item A training-free episode-level retry selector consistently improves multiple VLA backbones under retryable execution, with gains that closely track the recoverable headroom available to each policy.
\item Retrieval experiments show that the tested external demonstration prior is consistently useful for the backbone with the largest measured action-prior gap, while retrieval and selection provide additional gains when both opportunities are present.
\item Generalization experiments extend the analysis to a different robot and simulator, degraded observations, and an autoregressive VLA, revealing when available headroom transfers and when the ranking signal becomes the limiting factor.
\end{itemize}

\begin{figure}[t]
\centering
\includegraphics[width=0.98\linewidth]{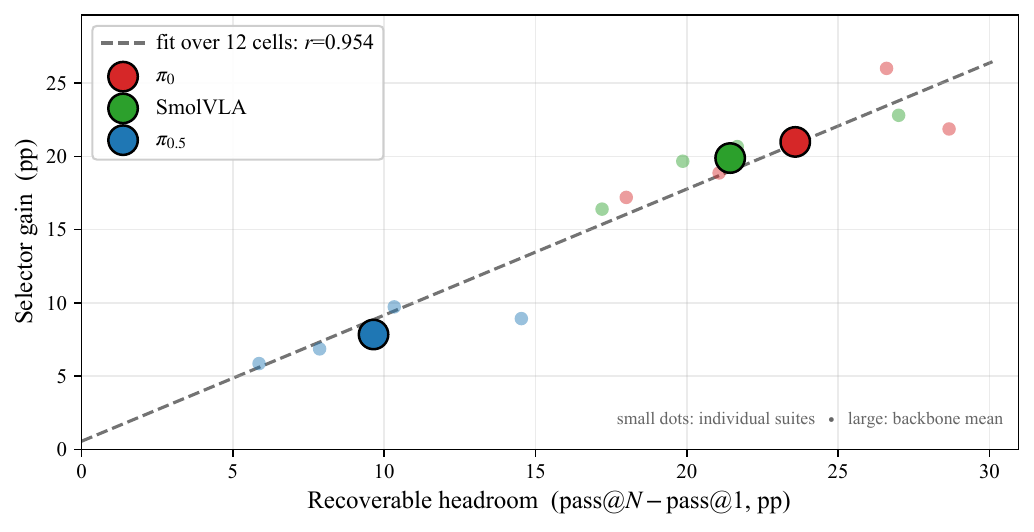}
\caption{\textbf{The headroom relation.} Selector gain against recoverable headroom (pass@$N-$pass@1). Small dots are the 12 backbone$\times$suite cells, and large markers are the three backbone means. The dashed fit gives $r{=}0.954$, bootstrap 95\% CI $[0.89,0.99]$, and $0.91{\pm}0.04$ on disjoint episode splits. The training-free score recovers about 90\% of the available headroom across backbones.}
\label{fig:law}
\end{figure}

\section{Related Work}
\noindent
\textbf{Retrieval-augmented robot policies.}
Non-parametric visual imitation and retrieval-based policy learning use nearby demonstrations, trajectories, or reusable sub-trajectories to improve robot behavior~\cite{pari2022surprising,du2023behavior,lin2024flowretrieval,memmel2024strap,kumar2025collage}. More recent methods incorporate retrieved experience directly into VLA action generation through demonstration-derived memory or trajectory context~\cite{li2025mapvla,park2026retrieve}. These works primarily study how external experience should be represented, retrieved, and incorporated into a policy. Our work instead asks when a frozen policy needs the retrieved prior in the first place. We measure how closely the policy already reproduces demonstrated actions and relate this mismatch to the observed benefit of retrieval across backbones.

\noindent
\textbf{Test-time compute and selection.}
Test-time scaling improves fixed models by generating multiple candidates and selecting among them~\cite{snell2024scaling}. Recent work extends this idea to robot policies and VLAs using action sampling, learned or vision-language-based verifiers, model-internal signals, and adaptive critics~\cite{kwok2025robomonkey,jang2025mgselect,dai2025rover,li2026vlaattc,zhao2026tapsampling}. Our work focuses on the opportunity available to selection rather than only the design of the selection rule. We measure recoverable headroom directly from repeated stochastic rollouts and use a training-free episode-level demonstration-manifold score to study how much of this existing capability can be recovered across policies and deployment conditions. Together with our retrieval analysis, this shifts the focus from applying a test-time intervention to diagnosing which form of augmentation a frozen policy actually needs.

\section{Method}
\subsection{Problem Setting}
\label{sec:setting}
We consider a frozen vision-language-action policy $\pi$ that produces a rollout $\tau \sim \pi(\cdot \mid s_0)$ from an initial episode state $s_0$. The policy parameters remain fixed for all test-time interventions considered in this work. Because the policy is stochastic, repeated executions from the same initial condition can produce different trajectories and task outcomes. Let $\mathrm{succ}(\tau) \in \{0,1\}$ denote whether a rollout successfully completes the task. We define single-rollout success and best-of-$N$ success as
\begin{equation}
\mathrm{pass@1}=\mathbb{E}_{\tau}[\mathrm{succ}(\tau)], \qquad
\mathrm{pass@}N=\mathbb{E}\left[\max_{i \leq N}\mathrm{succ}(\tau_i)\right].
\label{eq:passn}
\end{equation}
The difference between the two quantities defines the \textbf{recoverable headroom}
\begin{equation}
H_N=\mathrm{pass@}N-\mathrm{pass@1}.
\label{eq:headroom}
\end{equation}
Since a selector can only choose among the $N$ sampled rollouts, $H_N$ upper-bounds the improvement attainable by an oracle selector with access to $\mathrm{succ}(\cdot)$ over the same candidate set. On benchmarks with fixed initial states, we obtain the $N$ rollouts from $N$ policy seeds. The same rollouts are used to estimate $H_N$ and as the candidate set for the selector in Sec.~\ref{sec:selector}.

For retrieval, we separately consider \textbf{complementarity}. Rather than defining complementarity from downstream retrieval gains, we construct a probe that measures whether the demonstrated behavior is already represented by samples from the frozen policy.

\noindent
\textbf{Shared components.}
A demonstration library $\mathcal{D}=\{(o^{d}_j,a^{d}_j)\}_j$ contains observation frames and corresponding action chunks from successful task demonstrations. The selector uses the demonstration observations $\{o^d_j\}$, the demonstration-fit probe uses the actions $\{a^d_j\}$, and retrieval uses demonstration segments. A frozen $\ell_2$-normalized visual encoder $\phi$ maps observations to embeddings and is shared by the selector and retrieval. Its implementation is described in Sec.~\ref{sec:setup}.

\subsection{Measuring Retrieval Complementarity}
\label{sec:complementarity}
To measure how closely the policy's sampled action distribution matches demonstrated behavior, we evaluate the frozen policy on demonstration observations. Given a demonstration observation $o^d$ and its corresponding action chunk $a^\star$, we draw $K$ candidate chunks from the policy and define the demonstration-fit error as
\begin{equation}
\mathrm{DFE}=\mathbb{E}_{(o^d,a^\star)\sim\mathcal{D}}\left[\min_{k \leq K}\left\|a_k-a^\star\right\|_2\right],\qquad a_k\sim\pi(\cdot\mid o^d).
\label{eq:dfe}
\end{equation}
DFE measures the distance between a demonstrated action and the closest action available among the policy samples. In our LIBERO analysis, all three backbones operate in the same action space, and policy samples and demonstrated actions are compared using the same action representation. Lower values therefore indicate greater overlap between the sampled policy distribution and demonstrated behavior under this shared representation, while higher values indicate a larger mismatch. We use the same $K$ and demonstration states across backbones. Because the absolute scale of DFE depends on the action representation and sampling budget, we interpret it comparatively across policies evaluated under the same protocol rather than as a universally calibrated quantity. The probe is computed independently of downstream retrieval performance.

\subsection{Episode-Level Retry Selection}
\label{sec:selector}
\begin{figure}[t]
\centering
\includegraphics[width=0.99\linewidth]{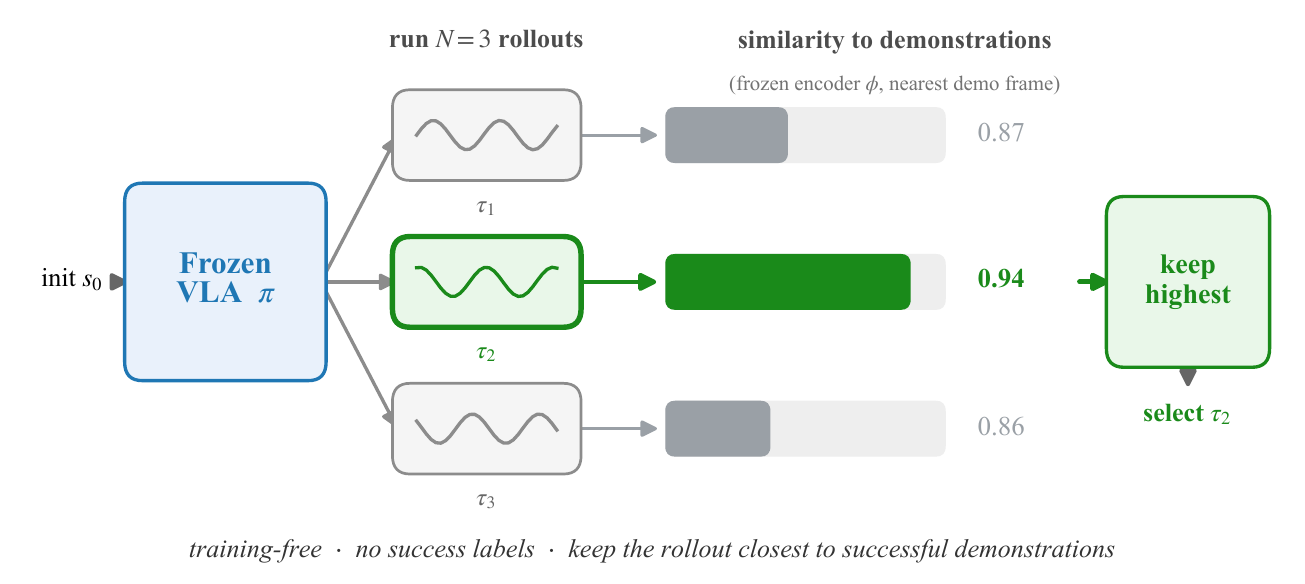}
\caption{\textbf{Episode-level retry selection.} From the same initial state $s_0$, the frozen policy $\pi$ is run $N$ times with different seeds, producing candidate rollouts $\tau_1,\dots,\tau_N$. Each rollout is scored by its average proximity to successful demonstrations in a frozen embedding space $\phi$ (Eq.~\ref{eq:manifold_score}), and the highest-scoring rollout is selected. The selector is training-free, uses no success labels, and assumes repeated or parallel execution.}
\label{fig:method}
\end{figure}

For a given episode, we execute the frozen policy $N$ times from the same initial condition, producing candidate trajectories $\{\tau_i\}_{i=1}^{N}$. Each trajectory contains a sequence of observations $\{o_t\}_{t=1}^{T}$. Using the shared encoder $\phi$ and demonstration frames $\{o^d_j\}$, we score each trajectory by its average proximity to the demonstration manifold
\begin{equation}
s(\tau)=\frac{1}{T}\sum_{t=1}^{T}\max_j \left\langle \phi(o_t),\phi(o^d_j)\right\rangle.
\label{eq:manifold_score}
\end{equation}
Because $\phi$ is $\ell_2$-normalized, the inner product corresponds to cosine similarity. For each rollout observation, we find the most similar demonstration frame in embedding space and average the similarity over the complete trajectory. We then select
\begin{equation}
\hat{\tau}=\arg\max_{i \leq N}s(\tau_i).
\label{eq:selector}
\end{equation}
Ties are broken by the lowest seed index. Selection is performed within each episode and requires no cross-task normalization. The score uses no success labels for the candidate rollouts and contains no trained scoring component. Figure~\ref{fig:method} summarizes the complete retry-selection pipeline.

\noindent
\textbf{Episode-level selection.}
We select complete rollouts rather than individual actions or action chunks. This matches the granularity at which task success is evaluated and avoids constructing a trajectory from independently selected local decisions. We compare episode-level and finer-grained selection in Sec.~\ref{sec:openvla}.

\noindent
\textbf{Execution setting and cost.}
The selector operates under retryable execution. The $N$ candidates are executed by resetting to a common initial state or through parallel execution on identical systems before one rollout is selected. The method therefore applies only to settings in which repeated or parallel executions are possible. Its execution cost is the standard $N\times$ overhead of running $N$ rollouts, while the scoring procedure itself requires only nearest-neighbor comparisons over frozen embeddings.

\subsection{Retrieval Augmentation}
\label{sec:retrieval}
We instantiate retrieval as event-schema prospective retrieval with three components. First, we identify event-start frames in the demonstration library using episode starts, gripper open or close transitions, and sharp changes in motion direction. These events define candidate sub-skill boundaries rather than arbitrary demonstration frames.

Second, the current event and each retrieval candidate are compared using a weighted sum of per-dimension-standardized distances over the frozen image embedding $\phi(o)$, the policy's predicted action chunk, and the proprioceptive state when available.

Third, retrieval is prospective. A candidate is matched using the current event but provides the action chunk associated with the next event in its demonstration. We denote this retrieved action chunk by $a_{\mathrm{retr}}$. For flow-matching policies, it is used to warm-start the sampler according to
\begin{equation}
a_{\mathrm{init}}=\alpha\,a_{\mathrm{retr}}+(1-\alpha)\,\epsilon,\qquad \alpha\in[0,1],
\label{eq:retrieval}
\end{equation}
where $\epsilon$ is the policy's standard sampling noise and $\alpha$ controls retrieval strength. The flow-matching ODE is then integrated from $a_{\mathrm{init}}$.

We additionally evaluate retrieval variants based on onset alignment, feature whitening, re-ranking, gripper-phase gating, reliability weighting, and variable lead-time retrieval. The main experiments use the strongest configuration selected using the held-out criterion described in the supplementary material. The exact event thresholds, retrieval weights, and selection protocol are also reported there for reproducibility. The selected configuration and hyperparameters are kept fixed across the reported LIBERO backbone comparisons. The training-free designation in this work refers specifically to the episode-level selector.

\section{Experimental Setup}
\label{sec:setup}
\noindent
\textbf{Backbones.}
We evaluate three frozen flow-matching VLA backbones finetuned on LIBERO, $\pi_0$~\cite{black2024pi0}, SmolVLA~\cite{shukor2025smolvla}, and $\pi_{0.5}$~\cite{intelligence2025pi05}. These models provide different capability and per-suite performance profiles while sharing the same LIBERO action space. We additionally evaluate the autoregressive OpenVLA-7B~\cite{kim2024openvla} to study whether the findings extend beyond flow-matching policies.

\noindent
\textbf{Benchmarks.}
Our primary evaluation uses the four LIBERO~\cite{liu2023libero} suites, Spatial, Object, Goal, and LIBERO-10, with the Franka robot. Each configuration is evaluated over $n{=}500{\times}3$ rollouts. The three stochastic seeds are used both to estimate recoverable headroom and as the $N{=}3$ candidates for episode-level selection.
We evaluate cross-simulator transfer on three SimplerEnv-Bridge~\cite{li2024simpler,walke2023bridgedata} tasks with the WidowX robot, carrot, spoon, and stack, using $n{=}24{\times}3$ rollouts. We further evaluate OpenVLA on the SimplerEnv coke-can task with the Google robot using 25 initial states and three sampled rollouts per state.
To study changes under degraded observations, we use LIBERO-Occ~\cite{li2026liberoocc}, an occlusion-oriented extension of LIBERO that introduces scene-induced visual occlusion into the manipulation environment. We evaluate the corresponding occluded task variants using $n{=}250{\times}2$ rollouts. We use the benchmark environments only and do not use its proposed Viewpoint Imagination method.

\noindent
\textbf{Implementation.}
The shared encoder $\phi$ concatenates frozen DINOv2~\cite{oquab2023dinov2} and SigLIP~\cite{zhai2023siglip} features into a 1,536-dimensional $\ell_2$-normalized representation. We use the wrist-camera view when available and otherwise use the available third-person view.
The demonstration library $\mathcal{D}$ contains the standard 50 LIBERO demonstrations per task~\cite{liu2023libero}. For the Bridge and coke-can evaluations, we construct task-specific demonstration libraries from successful demonstrations in the corresponding evaluation setting, containing 510 and 1,360 frames, respectively. The selector uses $N{=}3$ candidates for the main LIBERO evaluation. Retrieval uses $\alpha{=}0.3$ in Eq.~\ref{eq:retrieval}.
The demonstration-fit probe uses $K{=}3$ policy samples per demonstration state and is evaluated over 400 states. DFE comparisons across the three LIBERO backbones use the common action space and the same sampling protocol.
For the flow-matching evaluations, episodes run for up to 400 environment steps with 50-step action chunks. Preliminary evaluations using $n{=}100$ episodes occasionally produced changes of up to approximately four success-rate points that did not persist under the full protocol. We therefore report the larger evaluation protocols described above.

\section{Results}
\label{sec:results}
\subsection{Selection and retrieval expose different opportunities}
\label{sec:hierarchy}
Our main comparison reveals a clear difference in how selection and retrieval behave across frozen policies. Table~\ref{tab:main} shows that the episode-level selector improves every backbone and all 12 backbone$\times$suite cells, with average gains ranging from $+7.8$ to $+21.0$. Retrieval is more selective, providing positive gains across all four suites only for $\pi_0$. This difference cannot be explained by baseline capability alone. $\pi_0$ and SmolVLA have similar mean pass@1 performance, 64.2 and 65.6, yet the same retrieval mechanism consistently improves $\pi_0$ and has little effect on SmolVLA. Retrieval also provides only small and inconsistent changes for $\pi_{0.5}$.

\begin{table}[t]
\centering
\caption{\textbf{The episode-level selector improves every tested backbone, while retrieval provides consistent gains only for $\pi_0$.} LIBERO 4-suite mean success rate (\%) with gains over the frozen pass@1 baseline in parentheses. Selector gains are positive in all 12 backbone$\times$suite cells, while retrieval is positive across all four suites only for $\pi_0$.}
\label{tab:main}
\begin{tabular}{lccc}
\toprule
Backbone & pass@1 (frozen) & $+$Retrieval & $+$Selector (ours) \\
\midrule
$\pi_0$ & 64.2 & 70.6 ($+6.4$) & \textbf{85.2 ($\mathbf{+21.0}$)} \\
SmolVLA & 65.6 & 65.9 ($+0.3$) & \textbf{85.5 ($\mathbf{+19.9}$)} \\
$\pi_{0.5}$ & 88.7 & 89.2 ($+0.5$) & \textbf{96.5 ($\mathbf{+7.8}$)} \\
\bottomrule
\end{tabular}
\end{table}

\begin{table}[t]
\centering
\caption{\textbf{Retrieval is backbone-specific rather than determined by baseline capability alone.} Retrieval gain over pass@1 across the four LIBERO suites, reported as mean$\pm$std over three seeds. The retrieval mechanism, demonstration library, and hyperparameters are fixed across backbones. Bold marks the only backbone with positive gains across all four suites.}
\label{tab:retr}
\begin{tabular}{lccccc}
\toprule
Backbone & Mean pass@1 & Spatial & Object & Goal & 10 \\
\midrule
$\pi_0$ & 64.2 & $\mathbf{+4.5{\pm}1.5}$ & $\mathbf{+9.8{\pm}2.5}$ & $\mathbf{+5.5{\pm}0.8}$ & $\mathbf{+6.0{\pm}2.3}$ \\
SmolVLA & 65.6 & $+0.5{\pm}0.9$ & $-0.2{\pm}2.0$ & $+1.6{\pm}0.7$ & $-0.9{\pm}1.6$ \\
$\pi_{0.5}$ & 88.7 & $-1.1{\pm}0.9$ & $+0.6{\pm}0.6$ & $+2.6{\pm}2.0$ & $-0.1{\pm}2.8$ \\
\bottomrule
\end{tabular}
\end{table}

\noindent
\textbf{Why retrieval helps $\pi_0$ consistently.}
The demonstration-fit error provides an independent diagnostic for this difference. Under the shared action representation and fixed sampling protocol, DFE is $0.042$ for $\pi_0$, $0.021$ for $\pi_{0.5}$, and $0.003$ for SmolVLA over 400 demonstration states. As a separate sanity check with a scale-independent correlation statistic, the closest sampled first action correlates with the demonstrated first action at $0.91$, $0.96$, and $0.98$ for $\pi_0$, $\pi_{0.5}$, and SmolVLA, respectively. Both measurements place $\pi_0$ farthest from the demonstrated action prior.

All three policies operate in the same LIBERO action space and are evaluated on the same demonstration states, allowing DFE to be compared across the tested backbones without relying on downstream task success. The probe also evaluates the policy on demonstration states rather than states reached after rollout errors, providing a conservative view of the mismatch that may arise during deployment. We therefore use DFE as a diagnostic of an action-prior gap within a shared action space and sampling protocol, rather than as a calibrated continuous predictor of retrieval gain.

The matched-capability comparison between $\pi_0$ and SmolVLA is particularly informative. Their mean pass@1 differs by only $1.4$ points, yet SmolVLA reproduces demonstrated actions much more closely and receives little benefit from retrieval. In contrast, $\pi_0$ shows the largest demonstration-fit error and improves across all four suites under the same retrieval mechanism. The retrieval variants evaluated in our sweep change individual performance values but preserve this qualitative backbone-specific pattern. Across the three tested LIBERO backbones, these results indicate that baseline success alone is insufficient for identifying retrieval opportunity, while agreement with the demonstrated action prior provides a complementary diagnostic.

\noindent
\textbf{Reuse and retrieval compose.}
Retrieval and selection operate on different sources of improvement. Retrieval modifies the candidate distribution by introducing an external action prior, while selection chooses among the stochastic executions produced by that distribution. To test whether the two opportunities can be exploited together, we apply the episode-level selector to $N{=}3$ retrieval-warm-started rollouts. Table~\ref{tab:compose} shows that the combined method reaches a mean gain of $+23.3$, compared with $+21.0$ from selection alone and $+6.4$ from retrieval alone. The combination therefore provides an additional $+2.4$ points over selection on average.

The additional benefit depends on how much opportunity remains after selection. On LIBERO-10, retrieval increases the pass@3 ceiling from $64.4$ to $71.0$, and the combined method improves by another $+6.2$ points over selection alone. On Goal, where selection already reaches 97.6\% success, the combination provides no further improvement. This pattern is consistent with the two interventions acting on different sources of opportunity. Retrieval can change which behaviors are available among the candidates, while selection determines which available behavior is retained.

\begin{table}[t]
\centering
\caption{\textbf{Retrieval and selection provide additional gains when combined on average.} $\pi_0$ on LIBERO with $n{=}500{\times}3$ rollouts. Values report success rate (\%) with gains over pass@1 in parentheses. The final column reports the additional gain over selection alone.}
\label{tab:compose}
\begin{tabular}{lccccc}
\toprule
Suite & pass@1 & $+$retrieval & $+$selector & retrieval$+$selector & combine $-$ selector \\
\midrule
spatial & 63.6 & 68.1 ($+4.5$) & 89.6 ($+26.0$) & 92.4 ($+28.8$) & $+2.8$ \\
object & 77.1 & 86.9 ($+9.8$) & 96.0 ($+18.9$) & 97.2 ($+20.1$) & $+1.2$ \\
goal & 80.4 & 85.9 ($+5.5$) & 97.6 ($+17.2$) & 96.8 ($+16.4$) & $-0.8$ \\
10 & 35.7 & 41.7 ($+6.0$) & 57.6 ($+21.9$) & \textbf{63.8 ($+28.1$)} & $+6.2$ \\
\midrule
mean & 64.2 & 70.6 ($+6.4$) & 85.2 ($+21.0$) & \textbf{87.5 ($+23.3$)} & $+2.4$ \\
\bottomrule
\end{tabular}
\end{table}

\subsection{The headroom relation}
\label{sec:law2}
\noindent
\textbf{Gain tracks headroom.}
Recoverable headroom measures the opportunity available within a fixed set of stochastic policy executions. A large $H_N$ means that repeated sampling already exposes successful behaviors beyond pass@1, while a small $H_N$ leaves little room for any selector operating on the same candidate set. The empirical question is therefore how much of this available opportunity the demonstration-manifold score can recover.

Across the 12 LIBERO backbone$\times$suite cells, the capture ratio $\mathrm{gain}/H_3$ has mean $0.90$, standard deviation $0.11$, and range $[0.61,1.00]$. Selector gain and recoverable headroom have correlation $r{=}0.954$, with bootstrap 95\% confidence interval $[0.89,0.99]$ and Spearman correlation $\rho{=}0.958$. The relationship spans policies with substantially different baseline success rates, indicating that headroom is more informative about the opportunity available to selection than pass@1 alone.

\noindent
\textbf{The relation is not a shared-sample artifact.}
Because headroom and selector gain both contain the pass@1 term, we repeat the analysis using disjoint episodes. Headroom is measured on one split and selector gain on another, with no shared episodes. Across 50 random splits, the correlation remains $r{=}0.91{\pm}0.04$. Within individual backbones, correlations across the four suites are $0.81$, $0.95$, and $0.79$, while leave-one-backbone-out evaluation gives values between $0.86$ and $0.96$. These results support a stable empirical relationship between available headroom and selector gain rather than a correlation induced only by shared evaluation samples.

\noindent
\textbf{Selection, not resampling.}
Table~\ref{tab:ablation} separates the effect of generating additional rollouts from the effect of selecting among them. Randomly retaining one of three rollouts remains near pass@1, while demonstration-manifold selection improves all three backbones. Performance also increases from selection with $N{=}2$ to $N{=}3$, showing that additional samples become useful when accompanied by a ranking mechanism that can distinguish better executions.

We additionally evaluate the aggregate association between the ranking score and task success by pooling rollout-level manifold scores across all LIBERO backbones and suites. Using task success as the positive label gives an AUC of $0.96$ over $17{,}964$ valid rollouts. Of the $18{,}000$ LIBERO rollouts, $36$ do not have a valid manifold score and are excluded from this analysis. This pooled AUC indicates a strong aggregate association between manifold proximity and success, but it does not directly measure ranking among candidates from the same initial state. We therefore treat it as complementary evidence rather than as a direct measure of the within-episode selection problem. The end-to-end selector gains in Table~\ref{tab:ablation} provide the direct evaluation of whether the score can recover better outcomes from the available candidate sets.

\begin{table}[t]
\centering
\caption{\textbf{The gain comes from selecting additional samples rather than sampling alone.} LIBERO 4-suite mean success rate (\%). Randomly retaining one of three rollouts remains near pass@1, while manifold scoring recovers a large fraction of the available pass@3 headroom.}
\label{tab:ablation}
\begin{tabular}{lccccc}
\toprule
Backbone & pass@1 & rand-of-3 & selector ($N{=}2$) & selector ($N{=}3$) & oracle pass@3 \\
\midrule
$\pi_0$ & 64.2 & 64.5 & 79.2 & 85.2 & 87.8 \\
SmolVLA & 65.6 & 65.3 & 81.0 & 85.5 & 87.0 \\
$\pi_{0.5}$ & 88.7 & 89.0 & 94.8 & 96.5 & 98.3 \\
\bottomrule
\end{tabular}
\end{table}

\noindent
\textbf{Headroom and retrieval complementarity capture different opportunities.}
Recoverable headroom does not show the same relationship with retrieval gain, with a correlation of $r{=}0.32$. Retrieval is close to zero in eight of the twelve backbone$\times$suite cells despite substantial selector headroom in several of them. The contrast between $\pi_0$ and SmolVLA is again informative. Both have similar baseline performance and substantial recoverable headroom, but only $\pi_0$ exhibits a large demonstration-fit error and consistently benefits from retrieval.

These results distinguish two sources of test-time improvement. Headroom measures whether better outcomes are already present among repeated executions of the frozen policy, while demonstration fit measures how closely the policy reproduces the demonstrated action prior under a shared action representation. A policy may therefore have substantial headroom without benefiting from the tested retrieval mechanism. Conversely, retrieval may alter the candidate distribution without determining whether the resulting candidates can be reliably selected. We interpret DFE as evidence of retrieval complementarity within the tested LIBERO backbones rather than as a universal predictor of retrieval gain across arbitrary policies and action spaces.

\subsection{Generalization across shifts}
\label{sec:generalization}
\noindent
\textbf{Cross-simulator transfer.}
We next evaluate $\pi_0$ finetuned on BridgeData~\cite{walke2023bridgedata} tasks in SimplerEnv. Table~\ref{tab:bridge} shows that retrieval does not transfer on average, producing a $-0.9$ point change. In contrast, the same episode-level selector improves overall success by $+12.5$ points against an oracle headroom of $+18.1$, recovering 69\% of the available opportunity. This result extends the selector beyond the LIBERO robot, simulator, and demonstration library while also showing that retrieval behavior can change across domains.

\begin{table}[t]
\centering
\caption{\textbf{The selector transfers to a different robot and simulator, while retrieval does not improve overall performance.} SimplerEnv-Bridge with $\pi_0$-Bridge and WidowX using $n{=}24{\times}3$ rollouts. Values report success rate (\%) with gains over pass@1 in parentheses, and the oracle reports pass@$N$.}
\label{tab:bridge}
\begin{tabular}{lcccc}
\toprule
Task & pass@1 & $+$retrieval & $+$selector (ours) & oracle pass@$N$ \\
\midrule
carrot & 34.7 & 30.6 ($-4.2$) & \textbf{41.7 ($+6.9$)} & 50.0 \\
spoon & 45.8 & 47.2 ($+1.4$) & \textbf{54.2 ($+8.3$)} & 62.5 \\
stack & 31.9 & 31.9 ($-0.0$) & \textbf{54.2 ($+22.2$)} & 54.2 \\
\midrule
overall & 37.5 & 36.6 ($-0.9$) & \textbf{50.0 ($\mathbf{+12.5}$)} & 55.6 \\
\bottomrule
\end{tabular}
\end{table}

\noindent
\textbf{Cross-architecture evaluation.}
\label{sec:openvla}
We next test how the distinction between available headroom and ranking ability extends to an autoregressive VLA. On the SimplerEnv coke-can task, stochastic OpenVLA sampling alone leaves success unchanged at $0.40$, while independently re-ranking actions at each step decreases success to $0.36$. Episode-level selection reaches $0.44$ against an oracle pass@3 of $0.72$, as shown in Table~\ref{tab:openvla}. The main observation in this experiment is not the absolute $+0.04$ gain, which is noisy under the $n{=}25$ evaluation, but the large gap between realized selection and the oracle candidate set. The oracle pass@3 of $0.72$ shows that substantially better behavior is present among the sampled candidates, while the episode-level selector recovers only a small fraction of that headroom. The OpenVLA evaluation differs from the LIBERO setting in several respects, including the available observation views, and we therefore do not attribute the weaker headroom capture to any single factor. Instead, the result illustrates that substantial recoverable headroom does not necessarily translate into a large realized selection gain when the available ranking signal is insufficient to identify successful rollouts.

\begin{table}[t]
\centering
\caption{\textbf{On an autoregressive VLA, available headroom and realized selection can differ substantially.} OpenVLA-7B on the SimplerEnv coke-can task using 25 initial states. Gains are relative to greedy decoding, and oracle pass@3 indicates whether any of the three sampled episodes succeeds.}
\label{tab:openvla}
\begin{tabular}{llc}
\toprule
& Configuration & Success \\
\midrule
baseline & greedy decoding & 0.40 \\
& sample@1 ($T{=}1$, 3-seed mean) & 0.40 ($+0.00$) \\
per-step & re-rank ($N{=}4$) & 0.36 ($-0.04$) \\
episode-level (ours) & selector ($N{=}3$) & \textbf{0.44 ($+0.04$)} \\
\midrule
& oracle pass@3 & 0.72 \\
\bottomrule
\end{tabular}
\end{table}

\noindent
\textbf{Cross-condition evaluation.}
We finally test whether the available opportunity changes when the same policy is evaluated under scene-induced occlusion using LIBERO-Occ~\cite{li2026liberoocc}. Under this degraded observation condition, the performance of $\pi_{0.5}$ decreases substantially and its recoverable headroom increases. Its selector gain correspondingly increases from $+7.8$ under the clean condition to $+11.2$ under occlusion, approaching the $+11.9$ gain of $\pi_0$. Retrieval remains comparatively weak, with mean gains of $+4.3$ for $\pi_0$ and $+0.8$ for $\pi_{0.5}$.

These results show that the opportunity for test-time selection is not a fixed property of a backbone. Changing the observation condition changes the policy's stochastic success profile and therefore the amount of recoverable headroom available to selection. The increase in selector gain for $\pi_{0.5}$ under occlusion is consistent with this shift, while the comparatively weaker retrieval gains are consistent with the two interventions targeting different sources of opportunity.

\begin{table}[t]
\centering
\caption{\textbf{Under scene-induced occlusion, selector gains change together with the available opportunity.} LIBERO-Occ using $n{=}250{\times}2$ rollouts with $N{=}2$. Pass@1 reports the occluded baseline, while retrieval and selection columns report gains over it.}
\label{tab:occ}
\begin{tabular}{l|ccc|ccc}
\toprule
& \multicolumn{3}{c|}{$\pi_0$} & \multicolumn{3}{c}{$\pi_{0.5}$} \\
Suite & pass@1 & $+$retr & $+$sel & pass@1 & $+$retr & $+$sel \\
\midrule
spatial & 61.6 & $+6.2$ & $\mathbf{+18.4}$ & 80.8 & $+3.4$ & $\mathbf{+14.0}$ \\
object & 81.0 & $+6.6$ & $\mathbf{+9.4}$ & 74.0 & $+0.6$ & $\mathbf{+4.8}$ \\
goal & 74.4 & $+2.8$ & $\mathbf{+9.6}$ & 74.8 & $+2.0$ & $\mathbf{+12.0}$ \\
10 & 20.8 & $+1.8$ & $\mathbf{+10.4}$ & 42.4 & $-2.6$ & $\mathbf{+14.0}$ \\
\midrule
mean & 59.5 & $+4.3$ & $\mathbf{+11.9}$ & 68.0 & $+0.8$ & $\mathbf{+11.2}$ \\
\bottomrule
\end{tabular}
\end{table}

\section{Discussion}
\label{sec:discussion}
The results separate two questions that are often mixed together in test-time augmentation. Recoverable headroom measures whether additional successful behavior is already exposed by the policy's stochastic rollouts. Retrieval complementarity instead asks whether a retrieved action prior differs meaningfully from what the frozen policy already produces. These opportunities need not coincide. $\pi_0$ and SmolVLA have similar baseline performance and substantial headroom, yet only $\pi_0$ consistently benefits from retrieval. Conversely, large headroom does not guarantee a large realized selection gain when the ranking signal cannot reliably identify successful rollouts, as illustrated by the OpenVLA experiment.

These observations suggest a practical characterization of test-time augmentation opportunities rather than an equal-cost comparison between interventions. Repeated rollouts can be used to estimate whether recoverable headroom exists in a target condition when retryable or parallel execution is available. Selection can attempt to recover part of this existing capability without modifying the policy. Retrieval addresses a separate opportunity that can be probed by measuring how closely the frozen policy already reproduces a demonstrated action prior. Within the shared LIBERO action space used in our comparison, DFE identifies $\pi_0$ as the backbone with the largest measured action-prior gap. Retrieval and selection can also be combined when both forms of opportunity are present.

Our findings also refine the connection to test-time scaling in both general inference and VLA policies~\cite{snell2024scaling,kwok2025robomonkey,jang2025mgselect,dai2025rover,li2026vlaattc}. Additional samples are useful only when two conditions are satisfied. The candidate set must contain better outcomes, and the selection signal must identify them. Recoverable headroom measures the first condition, while the gap between oracle and realized selection exposes the second. The strong capture on LIBERO and the much weaker capture on OpenVLA show that these quantities should be treated separately rather than assuming that additional sampling alone will translate into better performance.

\section{Limitations}
\label{sec:limitations}
The main limitation of our approach is its execution model. Episode-level selection requires repeated or parallel executions and therefore applies only to settings where retries are possible. The reported gains should be interpreted as recoverable capability under retryable execution rather than as improvements to a single irreversible attempt. Selection and retrieval also incur different forms of test-time cost, and our experiments diagnose their distinct opportunities rather than establish an equal-budget comparison.

The effectiveness of selection depends on the trajectory-ranking signal, as illustrated by the lower headroom capture for OpenVLA. Because the OpenVLA evaluation differs from the LIBERO setting in several respects, including the available observation views, our experiments do not isolate the source of this weaker ranking performance. The reported global AUC measures an aggregate association with success rather than the exact within-episode ranking problem. DFE likewise depends on the action representation and sampling protocol and should not be interpreted as universally calibrated across policy families. Finally, our evidence for retrieval complementarity is limited to the tested backbones and retrieval mechanisms, and all experiments are conducted in simulation.

\bibliographystyle{splncs04}
\bibliography{main}

@article{black2024pi0,
  title={$\pi_0$: A Vision-Language-Action Flow Model for General Robot Control},
  author={Black, Kevin and Brown, Noah and Driess, Danny and Esmail, Adnan and Equi, Michael and Finn, Chelsea and Fusai, Niccolo and Groom, Lachy and Hausman, Karol and Ichter, Brian and others},
  journal={arXiv preprint arXiv:2410.24164},
  year={2024}
}

@article{intelligence2025pi05,
  title={$\pi_{0.5}$: A Vision-Language-Action Model with Open-World Generalization},
  author={{Physical Intelligence} and Black, Kevin and Brown, Noah and Darpinian, James and Dhabalia, Karan and Driess, Danny and Esmail, Adnan and Equi, Michael and Finn, Chelsea and Fusai, Niccolo and others},
  journal={arXiv preprint arXiv:2504.16054},
  year={2025}
}

@article{shukor2025smolvla,
  title={SmolVLA: A Vision-Language-Action Model for Affordable and Efficient Robotics},
  author={Shukor, Mustafa and Aubakirova, Dana and Capuano, Francesco and Kooijmans, Pepijn and Palma, Steven and Zouitine, Adil and Aractingi, Michel and Pascal, Caroline and Russi, Martino and Marafioti, Andres and others},
  journal={arXiv preprint arXiv:2506.01844},
  year={2025}
}

@inproceedings{kim2024openvla,
  title={OpenVLA: An Open-Source Vision-Language-Action Model},
  author={Kim, Moo Jin and Pertsch, Karl and Karamcheti, Siddharth and Xiao, Ted and Balakrishna, Ashwin and Nair, Suraj and Rafailov, Rafael and Foster, Ethan and Lam, Grace and Sanketi, Pannag and others},
  booktitle={Conference on Robot Learning},
  year={2024}
}

@inproceedings{liu2023libero,
  title={LIBERO: Benchmarking Knowledge Transfer for Lifelong Robot Learning},
  author={Liu, Bo and Zhu, Yifeng and Gao, Chongkai and Feng, Yihao and Liu, Qiang and Zhu, Yuke and Stone, Peter},
  booktitle={Advances in Neural Information Processing Systems},
  year={2023}
}

@inproceedings{li2024simpler,
  title={Evaluating Real-World Robot Manipulation Policies in Simulation},
  author={Li, Xuanlin and Hsu, Kyle and Gu, Jiayuan and Pertsch, Karl and Mees, Oier and Walke, Homer Rich and Fu, Chuyuan and Lunawat, Ishikaa and Sieh, Isabel and Kirmani, Sean and others},
  booktitle={Conference on Robot Learning},
  year={2024}
}

@inproceedings{walke2023bridgedata,
  title={BridgeData V2: A Dataset for Robot Learning at Scale},
  author={Walke, Homer Rich and Black, Kevin and Zhao, Tony Z. and Vuong, Quan and Zheng, Chongyi and Hansen-Estruch, Philippe and He, Andre Wang and Myers, Vivek and Kim, Moo Jin and Du, Max and others},
  booktitle={Conference on Robot Learning},
  year={2023}
}

@inproceedings{memmel2024strap,
  title={STRAP: Robot Sub-Trajectory Retrieval for Augmented Policy Learning},
  author={Memmel, Marius and Berg, Jacob and Chen, Bingqing and Gupta, Abhishek and Francis, Jonathan},
  booktitle={International Conference on Learning Representations},
  year={2025}
}

@inproceedings{du2023behavior,
  title={Behavior Retrieval: Few-Shot Imitation Learning by Querying Unlabeled Datasets},
  author={Du, Maximilian and Nair, Suraj and Sadigh, Dorsa and Finn, Chelsea},
  booktitle={Robotics: Science and Systems},
  year={2023}
}

@article{snell2024scaling,
  title={Scaling LLM Test-Time Compute Optimally Can Be More Effective than Scaling Model Parameters},
  author={Snell, Charlie and Lee, Jaehoon and Xu, Kelvin and Kumar, Aviral},
  journal={arXiv preprint arXiv:2408.03314},
  year={2024}
}

@article{kwok2025robomonkey,
  title={RoboMonkey: Scaling Test-Time Sampling and Verification for Vision-Language-Action Models},
  author={Kwok, Jacky and Agia, Christopher and Sinha, Rohan and Foutter, Matt and Li, Shulu and Stoica, Ion and Mirhoseini, Azalia and Pavone, Marco},
  journal={arXiv preprint arXiv:2506.17811},
  year={2025}
}

@article{oquab2023dinov2,
  title={DINOv2: Learning Robust Visual Features without Supervision},
  author={Oquab, Maxime and Darcet, Timoth{\'e}e and Moutakanni, Th{\'e}o and Vo, Huy and Szafraniec, Marc and Khalidov, Vasil and Fernandez, Pierre and Haziza, Daniel and Massa, Francisco and El-Nouby, Alaaeldin and others},
  journal={Transactions on Machine Learning Research},
  year={2024}
}

@inproceedings{zhai2023siglip,
  title={Sigmoid Loss for Language Image Pre-Training},
  author={Zhai, Xiaohua and Mustafa, Basil and Kolesnikov, Alexander and Beyer, Lucas},
  booktitle={International Conference on Computer Vision},
  year={2023}
}

@inproceedings{lin2024flowretrieval,
  title={FlowRetrieval: Flow-Guided Data Retrieval for Few-Shot Imitation Learning},
  author={Lin, Li-Heng and Cui, Yuchen and Xie, Amber and Hua, Tianyu and Sadigh, Dorsa},
  booktitle={Proceedings of the 8th Conference on Robot Learning},
  pages={4084--4099},
  volume={270},
  series={Proceedings of Machine Learning Research},
  publisher={PMLR},
  year={2025}
}

@inproceedings{kumar2025collage,
  title={COLLAGE: Adaptive Fusion-Based Retrieval for Augmented Policy Learning},
  author={Kumar, Sateesh and Dass, Shivin and Pavlakos, Georgios and Mart{\'i}n-Mart{\'i}n, Roberto},
  booktitle={Proceedings of the 9th Conference on Robot Learning},
  pages={4607--4624},
  volume={305},
  series={Proceedings of Machine Learning Research},
  publisher={PMLR},
  year={2025}
}

@inproceedings{jang2025mgselect,
  title={Verifier-Free Test-Time Sampling for Vision Language Action Models},
  author={Jang, Suhyeok and Kim, Dongyoung and Kim, Changyeon and Kim, Youngsuk and Shin, Jinwoo},
  booktitle={International Conference on Learning Representations},
  year={2026}
}

@article{dai2025rover,
  title={RoVer: Robot Reward Model as Test-Time Verifier for Vision-Language-Action Model},
  author={Dai, Mingtong and Liu, Lingbo and Bai, Yongjie and Liu, Yang and Wang, Zhouxia and Su, Rui and Chen, Chunjie and Lin, Liang and Wu, Xinyu},
  journal={arXiv preprint arXiv:2510.10975},
  year={2025}
}

@inproceedings{pari2022surprising,
  title={The Surprising Effectiveness of Representation Learning for Visual Imitation},
  author={Pari, Jyothish and Shafiullah, Nur Muhammad and Arunachalam, Sridhar Pandian and Pinto, Lerrel},
  booktitle={Proceedings of Robotics: Science and Systems},
  address={New York City, NY, USA},
  month={June},
  doi={10.15607/RSS.2022.XVIII.010},
  year={2022}
}

@article{park2026retrieve,
  title={Retrieve, Don't Retrain: Extending Vision Language Action Models to New Tasks at Test Time},
  author={Park, Jeongeun and Park, Juhan and Kim, Taekyung and Choi, Sungjoon and Han, Dongyoon and Yun, Sangdoo},
  journal={arXiv preprint arXiv:2606.15631},
  year={2026}
}

@article{li2026vlaattc,
  title={VLA-ATTC: Adaptive Test-Time Compute for VLA Models with Relative Action Critic Model},
  author={Li, Wenhao and Su, Xiu and Niu, Dan and Cao, Yichao and Xu, Hongyan and Qu, Zhe and Fan, Lei and You, Shan and Xu, Chang},
  journal={arXiv preprint arXiv:2605.01194},
  year={2026}
}

@article{li2025mapvla,
  title={MAP-VLA: Memory-Augmented Prompting for Vision-Language-Action Model in Robotic Manipulation},
  author={Li, Runhao and Guo, Wenkai and Wu, Zhenyu and Wang, Changyuan and Deng, Haoyuan and Weng, Zhenyu and Tan, Yap-Peng and Wang, Ziwei},
  journal={arXiv preprint arXiv:2511.09516},
  year={2025}
}

@inproceedings{zhao2026tapsampling,
  title={TapSampling: Inference-Time Sampling with a Task-Progress-Understanding Verifier for Robotic Manipulation},
  author={Zhao, Sizhe and Zhang, Shengping and Yang, Shuo and Zhao, Weiyu and Wang, Shuigen and Ji, Xiangyang},
  booktitle={International Conference on Machine Learning},
  year={2026}
}

@article{li2026liberoocc,
  title={{LIBERO-Occ}: Evaluating and Improving Vision-Language-Action Models under Scene-Induced Occlusion via Viewpoint Imagination},
  author={Li, Taishan and Zhang, Jiwen and Wang, Siyuan and Huang, Xuanjing and Wei, Zhongyu},
  journal={arXiv preprint arXiv:2606.10862},
  year={2026}
}

\clearpage
\section*{Supplementary Material}
This supplement provides the implementation details referenced in the main paper: the
retrieval variants we evaluated, how the strongest configuration was selected on held-out
data, the event-detection rule, the retrieval distance, and a summary of all fixed
hyperparameters. All settings are shared across backbones unless stated otherwise.

\setcounter{section}{0}
\renewcommand{\thesection}{S\arabic{section}}
\setcounter{table}{0}
\renewcommand{\thetable}{S\arabic{table}}

\section{Retrieval Variants and Configuration Selection}
\label{sec:supp_retr}
The main retrieval intervention is event-schema prospective retrieval (main paper,
Sec.~3.4). Around this core mechanism we evaluated the following variants:
\begin{itemize}
\item \textbf{Onset alignment}: aligning the query to the onset frame of the current
event rather than the current frame.
\item \textbf{Feature whitening}: per-dimension whitening of the frozen embedding before
computing retrieval distances.
\item \textbf{Re-ranking}: retrieving a short candidate list and re-ranking it with the
combined image--action--state distance.
\item \textbf{Gripper-phase gating}: restricting candidates to demonstration segments in
the same gripper phase (open/closed) as the query.
\item \textbf{Reliability weighting}: down-weighting candidates whose retrieved action
chunk disagrees with the policy's own prediction.
\item \textbf{Variable lead-time retrieval}: varying how many events ahead the retrieved
action chunk is taken from.
\end{itemize}

\noindent\textbf{Held-out selection.}
We do not tune these choices on the reported evaluation episodes, and we do
\emph{not} tune them per backbone. The retrieval variant and the retrieval strength
$\alpha$ are selected \emph{once}, on a disjoint held-out split of demonstrations and
initial states, and the resulting single configuration is then \emph{frozen} and applied
unchanged to \emph{every} backbone and every reported episode. Using one common
configuration and a single $\alpha{=}0.3$ across all backbones is exactly what makes the
backbone-specific retrieval pattern in the main paper (Table~2) a controlled comparison
rather than an artifact of per-backbone tuning. Individual variants change absolute
performance values but preserve this qualitative ordering (retrieval helps the backbone
with the largest demonstration-fit error).

\section{Event Detection}
\label{sec:supp_events}
Event-start frames partition each demonstration into sub-skills. A frame is marked as an
event start if any of the following holds: (i) it is the first frame of the episode;
(ii) the gripper state transitions between open and closed; or (iii) the end-effector
motion direction changes by more than a fixed angular threshold between consecutive steps.
The gripper-state and motion-direction thresholds are set once on the held-out split
(Sec.~\ref{sec:supp_retr}) and then kept fixed for all backbones and suites. Prospective
retrieval matches the current event but returns the action chunk associated with the
\emph{next} event in the retrieved demonstration.

\section{Retrieval Distance}
\label{sec:supp_dist}
A query event and a candidate event are compared with a weighted sum of per-dimension
standardized (z-scored) distances over three modalities: the frozen image embedding
$\phi(o)$ (concatenated DINOv2 and SigLIP features, $1{,}536$-d, $\ell_2$-normalized), the
policy's predicted action chunk, and the proprioceptive state when available. Each modality
is standardized independently so that the three contribute on comparable scales; we use
equal modality weights by default. The retrieved chunk warm-starts the flow-matching
sampler as $a_{\mathrm{init}}=\alpha\,a_{\mathrm{retr}}+(1-\alpha)\,\epsilon$ with
$\alpha{=}0.3$.

\section{Fixed Hyperparameters}
\label{sec:supp_hparams}
Table~\ref{tab:supp_hp} summarizes the settings used throughout. The same
demonstration-manifold score, encoder, and candidate count are used for every backbone
without backbone-specific training.

\begin{table}[h]
\centering
\caption{Fixed hyperparameters shared across backbones and suites.}
\label{tab:supp_hp}
\begin{tabular}{@{}p{4.3cm}p{7.2cm}@{}}
\toprule
Setting & Value \\
\midrule
Visual encoder $\phi$ & frozen DINOv2 $+$ SigLIP, $1{,}536$-d, $\ell_2$-normalized \\
Camera view for scoring & wrist camera when available, else third-person \\
Selector candidates $N$ & $3$, from $3$ seeds (LIBERO) \\
Demonstration library & $50$ demos/task (LIBERO); $510$/$1{,}360$ frames (Bridge/coke-can) \\
Retrieval strength $\alpha$ & $0.3$ \\
Demonstration-fit samples $K$ & $3$, over $400$ demonstration states \\
Episode length & up to $400$ steps, $50$-step action chunks \\
Evaluation protocol & LIBERO $n{=}500{\times}3$; Bridge $24{\times}3$; coke-can $25$; LIBERO-Occ $250{\times}2$ \\
\bottomrule
\end{tabular}
\end{table}

\noindent
The training-free designation in the main paper refers specifically to the episode-level
selector, which uses only frozen embeddings and nearest-neighbor comparisons to
demonstrations and requires no task-success labels or additional training.

\end{document}